\documentclass[nofootinbib,superscriptaddress,a4paper,twocolumn,longbibliography]{revtex4-1}

\usepackage{graphicx}
\usepackage{dcolumn}
\usepackage{bm}
\usepackage{subfigure}
\usepackage{physics}
\usepackage{tikz}
\usepackage{qcircuit}
\usepackage{booktabs}
\usepackage{adjustbox}

\usepackage{hyperref}
\usepackage{tikz}
\usepackage{calc}

\newcommand\blfootnote[1]{%
  \begingroup
  \renewcommand\thefootnote{}\footnote{#1}%
  \addtocounter{footnote}{-1}%
  \endgroup
}

\usepackage{hyperref}
\usepackage{url}
\usepackage{graphicx}
\usepackage{xcolor}
\usepackage{xspace}
\usepackage{wrapfig}
\usepackage{booktabs}

\usepackage{amssymb}
\usepackage{pifont}

\graphicspath{
	{pics/}
}
\begin{document}

\title{Assigning Confidence to Molecular Property Prediction}
\affiliation{Department of Computer Science, University of Toronto, Canada.} 
\affiliation{Department of Chemistry, University of Toronto, Canada.}

\author{AkshatKumar Nigam}
\affiliation{Department of Computer Science, University of Toronto, Canada.} 
\affiliation{Department of Chemistry, University of Toronto, Canada.}

\author{Robert Pollice}
\affiliation{Department of Computer Science, University of Toronto, Canada.} 
\affiliation{Department of Chemistry, University of Toronto, Canada.}

\author{Matthew F. D. Hurley}
\affiliation{Department of Computer Science, University of Toronto, Canada.} 
\affiliation{Department of Chemistry, Temple University, Philadelphia, PA 19122, USA.}

\author{Riley J. Hickman}
\affiliation{Department of Computer Science, University of Toronto, Canada.} 
\affiliation{Department of Chemistry, University of Toronto, Canada.}

\author{Matteo Aldeghi}
\affiliation{Department of Computer Science, University of Toronto, Canada.} 
\affiliation{Department of Chemistry, University of Toronto, Canada.}
\affiliation{Vector Institute for Artificial Intelligence, 661 University Ave Suite 710, Toronto, Ontario M5G 1M1, Canada.}

\author{Naruki Yoshikawa}
\affiliation{Department of Computer Science, University of Toronto, Canada.} 
\affiliation{Department of Chemistry, University of Toronto, Canada.}

\author{Seyone Chithrananda}
\affiliation{Department of Computer Science, University of Toronto, Canada.} 

\author{Vincent A. Voelz}
\affiliation{Department of Computer Science, University of Toronto, Canada.} 
\affiliation{Department of Chemistry, Temple University, Philadelphia, PA 19122, USA.}

\author{Al\'an Aspuru-Guzik}
\email{Correspondence to: alan@aspuru.com}
\affiliation{Department of Computer Science, University of Toronto, Canada.}
\affiliation{Department of Chemistry, University of Toronto, Canada.}
\affiliation{Vector Institute for Artificial Intelligence, Toronto, Canada.}
\affiliation{Lebovic Fellow, Canadian Institute for Advanced Research (CIFAR), 661 University Ave, Toronto, Ontario M5G, Canada. }

\begin{abstract}
\textbf{Introduction:} Computational modeling has rapidly advanced over the last decades, especially to predict molecular properties for chemistry, material science and drug design. Recently, machine learning techniques have emerged as a powerful and cost-effective strategy to learn from existing datasets and perform predictions on unseen molecules. Accordingly, the explosive rise of data-driven techniques raises an important question: What confidence can be assigned to molecular property predictions and what techniques can be used for that purpose? \\ \textbf{Areas covered: } In this work, we discuss popular strategies for predicting molecular properties relevant to drug design, their corresponding uncertainty sources and methods to quantify uncertainty and confidence. First, our considerations for assessing confidence begin with dataset bias and size, data-driven property prediction and feature design. Next, we discuss property simulation via molecular docking, and free-energy simulations of binding affinity in detail. Lastly, we investigate how these uncertainties propagate to generative models, as they are usually coupled with property predictors. \\ \textbf{Expert opinion: }Computational techniques are paramount to reduce the prohibitive cost and timing of brute-force experimentation when exploring the enormous chemical space. We believe that assessing uncertainty in property prediction models is essential whenever closed-loop drug design campaigns relying on high-throughput virtual screening are deployed. Accordingly, considering sources of uncertainty leads to better-informed experimental validations, more reliable predictions and to more realistic expectations of the entire workflow. Overall, this increases confidence in the predictions and designs and, ultimately, accelerates drug design. 
\end{abstract}

\maketitle 
\section{Introduction}\label{chp:intro}\blfootnote{Our references can be navigated:\\ \url{https://github.com/aspuru-guzik-group/assessing_mol_prediction_confidence}}
While data-driven modeling has made large advances in image processing and speech recognition, ground-breaking contributions in drug discovery are harder to find. Often, only limited amounts of reliable \textit{in vitro} and \textit{in vivo} data suitable for supervised learning tasks are available. Additionally, uncertainties of experimental data, for instance regarding effectiveness and toxicity, are common and significant as the experimental design is complex, prone to noise and the outcome is dependent on many parameters, in particular compound dosage. Hence, these uncertainties are propagated to property prediction workflows when data-driven models are trained on that data. \\

While it is evident that machine learning (ML) prediction accuracy and data quality are intrinsically related, there are more subtle aspects that are sometimes neglected. Dataset size, composition, coverage and training-test split have considerable influence on the final model performance but are insufficient to assess uncertainty alone without suitable models and methodologies for that purpose.  For instance, using random or scaffold-based train/test splits does not lead to reliable measures of model performance and generalizability. Other common problems include high dataset bias, small sample size or low chemical diversity. However, some of these issues currently do not have a straightforward remedy as there is no commonly accepted way to describe chemical subspaces rendering it difficult to uncover all inherent biases. Nevertheless, when the underlying data is not investigated appropriately model performance and generalizability tend to be overestimated leading to a significant number of false predictions and, ultimately, low confidence in the property prediction workflows. \\

In this article, we discuss methodologies and datasets for chemical properties important in drug design. In particular, we focus on the main sources of uncertainties for both simulation-based and data-driven property prediction. In doing so, we discuss uncertainties inherent in datasets, outputs of data-driven models, input features, and simulation of binding affinities. In addition, we discuss the importance of uncertainty in generative models, especially for property-based molecule design. Finally, we close with our personal opinion on the most important aspects of uncertainty and confidence focusing on important problems and future avenues that will lead to higher predictive ability and models that naturally take uncertainty into account.

\begin{figure*}[ht]
    \centering
    \includegraphics[width=0.75\textwidth]{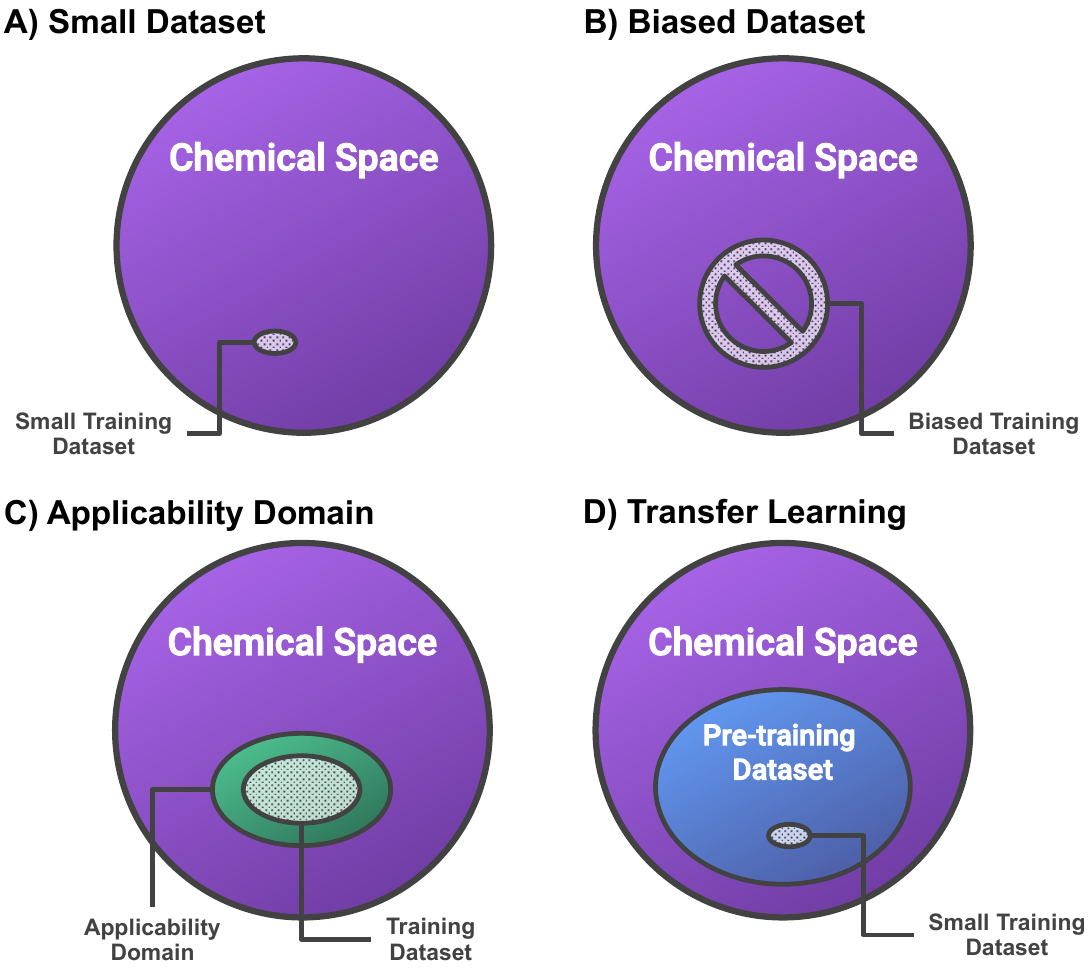}
    \caption{Common problems with datasets for data-driven molecular property prediction and strategies to improve confidence in the models. A) Small training datasets hamper generalization. B) Biased training datasets lead to models only learning the inherent bias rather than meaningful relationships. C) The applicability domain of a model defines the molecules for which the property predictions are expected to be reliable. D) Transfer learning via pre-training on large (unlabelled) datasets improves prediction accuracy for property prediction based on small training datasets.}
    \label{fig:fig2}
\label{fig_x}
\end{figure*}

\begin{table*}[]
\caption{Some of the common datasets used for molecular machine learning and data-driven property prediction.}
\begin{adjustbox}{width=0.98\textwidth}
\begin{tabular}{l l c l}
\hline
\textbf{Name}                           & \multicolumn{1}{c}{\textbf{Description}}                                                                                                                                                                                                                                 & \textbf{Number of molecules} & \multicolumn{1}{c}{\textbf{Possible bias}}                                                                                                                                                                                   \\ \hline
ZINC \cite{irwin2020zinc20}                           & \begin{tabular}[c]{@{}l@{}}Database of commercially available \\ compounds together with very simple \\ estimated molecular properties for\\  virtual screening.\end{tabular}                                                                                    & 1.4 billion         & \begin{tabular}[c]{@{}l@{}}Inherently biased by currently synthesizable \\ chemical space. Consequently, the molecular \\ shapes have been shown to be highly biased\\  against sphere-like molecules.\end{tabular} \\ \hline
QM9 \cite{ramakrishnan2014quantum}                        & \begin{tabular}[c]{@{}l@{}}Electronic properties estimated using \\ density functional theory (DFT) \\ simulations.\end{tabular}                                                                                                                                 & 134 thousand        & \begin{tabular}[c]{@{}l@{}}Biased towards small molecules only \\ containing  the elements C, H, N, O and F.\end{tabular}                                                                                           \\ \hline
PubChemQC \cite{nakata2017pubchemqc, nakata2020pubchemqc}                & \begin{tabular}[c]{@{}l@{}}Geometries and electronic properties \\ of molecules with short string \\ representations taken from PubChem.\end{tabular}                                                                                                            & 221 million         & \begin{tabular}[c]{@{}l@{}}Biased towards small molecules that \\ have been reported in the literature\\  before.\end{tabular}                                                                                      \\ \hline
Tox21 \cite{tox_data}                   & \begin{tabular}[c]{@{}l@{}}Toxicologic properties of molecules\\  with respect to 12 different assays\end{tabular}                                                                                                                                               & 13 thousand         & \begin{tabular}[c]{@{}l@{}}Biased towards environmental \\ compounds and approved drugs.\end{tabular}                                                                                                               \\ \hline
ToxCast \cite{tox_data_2}               & \begin{tabular}[c]{@{}l@{}}High-throughput screening and computational\\  data for the toxicology of molecules from \\ industry, consumer products and the food\\  industry based on cell assays.\end{tabular}                                                   & 1.8 thousand        & \begin{tabular}[c]{@{}l@{}}Biased towards molecules used\\  in industry, consumer products \\ and the food industry.\end{tabular}                                                                                   \\ \hline
ClinTox \cite{wu2018moleculenet}                       & \begin{tabular}[c]{@{}l@{}}Drugs and drug candidates that made\\  it to clinical trials and were either\\  approved or failed.\end{tabular}                                                                                                                      & 1.5 thousand        & \begin{tabular}[c]{@{}l@{}}Biased towards drugs that made\\ it to clinical trials.\end{tabular}                                                                                                                     \\ \hline
SIDER \cite{sider_}                         & \begin{tabular}[c]{@{}l@{}}Recorded adverse drug reactions\\  of marketed drugs.\end{tabular}                                                                                                                                                                    & 1.4 thousand        & Biased towards marketed drugs.                                                                                                                                                                                      \\ \hline
ChEMBL \cite{mendez2019chembl}                       & \begin{tabular}[c]{@{}l@{}}Bioactive small molecules and their \\ activities extracted from the literature, \\ from clinical trials and from other databases.\end{tabular}                                                                                       & 2.0 million         & \begin{tabular}[c]{@{}l@{}}Biased towards compounds for \\ which bioactivity was published \\ in the scientific literature.\end{tabular}                                                                            \\ \hline
DUD-E \cite{chen2019hidden}                        & \begin{tabular}[c]{@{}l@{}}Ligand binding affinities against 102 distinct\\  target proteins with both strong and weak \\ binders.\end{tabular}                                                                                                                  & 23 thousand         & \begin{tabular}[c]{@{}l@{}}Biased towards molecules that\\  have been synthesized and \\ evaluated for binding affinity.\end{tabular}                                                                               \\ \hline
AqSolDB \cite{keller2016olfactory}                       & \begin{tabular}[c]{@{}l@{}}Aqueous solubility data of organic molecules\\  taken from 9 different datasets.\end{tabular}                                                                                                                                         & 10 thousand         & \begin{tabular}[c]{@{}l@{}}Biased towards organic molecules\\ with relatively high aqueous \\ solubility.\end{tabular}                                                                                              \\ \hline
Olfaction Prediction Challenge \cite{keller2016olfactory} & \begin{tabular}[c]{@{}l@{}}Olfactory perception of organic molecules\\  at different concentrations.\end{tabular}                                                                                                                                                & 0.5 thousand        & \begin{tabular}[c]{@{}l@{}}Biased towards small and volatile \\ organic molecules. Results biased\\  by familiarity of smells.\end{tabular}                                                                         \\ \hline
FreeSolv \cite{mobley2014freesolv}                    & \begin{tabular}[c]{@{}l@{}}Experimental and computed hydration free\\  energies of small and neutral molecules.\end{tabular}                                                                                                                                     & 0.6 thousand        & \begin{tabular}[c]{@{}l@{}}Biased towards small and neutral \\ molecules that have been studied \\ in the literature both computationally\\  and experimentally for hydration \\ free energies.\end{tabular}        \\ \hline
ESOL \cite{delaney2004esol}                     & \begin{tabular}[c]{@{}l@{}}Experimental aqueous solubility combining\\  datasets for small molecules from the \\ literature, for medium-sized molecules \\ used as pesticides and larger proprietary \\ compounds from the pharmaceutical industry.\end{tabular} & 2.9 thousand        & \begin{tabular}[c]{@{}l@{}}The sub-groups each have a different\\  bias as they each have different \\ application domains.\end{tabular}                                                                            \\ \hline
Lipophilicity \cite{wang2015silico, li_2018}         & \begin{tabular}[c]{@{}l@{}}Experimental n-octanol/water (buffered at pH\\  7.4) distribution coefficient of organic \\ molecules taken from other databases.\end{tabular}                                                                                        & 1.1 thousand        & \begin{tabular}[c]{@{}l@{}}Biased towards molecules with \\ distribution coefficients between \\ -10 and 10.\end{tabular}                                                                                           \\ \hline
PubChem Bioassay \cite{wang2012pubchem}             & \begin{tabular}[c]{@{}l@{}}Bioactivity outcomes from high-throughput \\ screenings of molecules.\end{tabular}                                                                                                                                                    & 2.3 million         & \begin{tabular}[c]{@{}l@{}}Biased towards molecules of \\ interest and molecules that \\ are synthesizable.\end{tabular}                                                                                            \\ \hline
PDBbind \cite{wang2004pdbbind, liu2015pdb}                      & \begin{tabular}[c]{@{}l@{}}Experimental binding affinity for biomolecular \\ complexes deposited in the protein data \\ bank (PDB).\end{tabular}                                                                                                                 & 21.4 thousand       & \begin{tabular}[c]{@{}l@{}}Biased towards complexes with\\  available crystal structures.\end{tabular}                                                                                                              \\ \hline
BBBP \cite{martins2012bayesian}                   & \begin{tabular}[c]{@{}l@{}}The blood-brain penetration partition \\ coefficient for molecules collected from the \\ literature.\end{tabular}                                                                                                                     & 2.1 thousand        & \begin{tabular}[c]{@{}l@{}}Biased towards molecules \\ studied in the literature for \\ blood-brain penetration.\end{tabular}                                                                                       \\ \hline
\end{tabular}
\end{adjustbox}
\end{table*}

\section{Dataset uncertainty}\label{chp:method}
One of the most common challenges for data-driven molecular property prediction is dataset size and composition. For many important properties in drug discovery, especially pharmacologic properties of molecules such as absorption, distribution, metabolism, excretion and toxicity (ADMET), only a limited amount of high-quality data is available and it is usually only available for certain classes of molecules, which inherently introduces biases. To provide an overview of the amount and type of data available, Table 1 lists popular datasets for molecular property prediction. Accordingly, in this section, we discuss dataset characteristics that are important to consider for data-driven property prediction tasks along with methods and procedures to minimize their impact on property prediction performance. \\

In the framework of supervised learning, models learn relationships between input and output from the training data and, subsequently, predict properties for unlabelled input data, and the corresponding prediction ability is termed generalization. When predictors perform well on the training set, but performance on the unlabelled data is poor, the model is overfitted. This is a very common problem in ML, especially when the training set size is small (Figure 1-A). Learning theory states that collecting more data improves generalization if the data is independent and identically distributed \cite{goodfellow2016deep}. However, real data is not a uniform sample of chemical space. Typically, molecules in a dataset are collected under specific criteria such as the number of atoms, the constituent elements, the similarity to known molecules or the availability of synthetic procedures introducing bias (Figure 1-B). In addition, there is an inherent bias in both industry and academia to publish only successful experiments which can be problematic for assembling training data as negative results are as important as positive ones for data-driven modeling. \\

First, the uncertainty of model predictions for unlabelled molecules is largely dependent on the difference of the structure and property distributions to the training data. To quantify this difference, in the field of quantitative structure-activity relationships (QSARs) \cite{muratov2020qsar}, the concept of applicability domain (AD) is widely applied. The AD of a QSAR model is defined as “the response and chemical structure space in which the model makes predictions with a given reliability”\cite{netzeva2005current} and makes the inherent bias of a model comprehensible (Figure 1-C). There are several methods to estimate ADs \cite{pan2009survey}. A common approach is to consider molecules with descriptors within a certain distance to the mean of the training data to be inside the AD. Notably, predictions for molecules inside the AD are considered reliable, outside the AD reliability cannot be guaranteed. \\

Secondly, a significant bias in the training data can cause models only to learn the inherent bias rather than physically meaningful relationships (Figure 1-B). When users are unaware of this bias, it inadvertently leads to overly optimistic conclusions concerning model performance. Recently \cite{chen2019hidden}, it was pointed out that there is significant hidden bias in the widely used dataset for structure-based virtual screening, the Directory of Useful Decoys: Enhanced (DUD-E)\cite{mysinger2012directory}. DUD-E includes two types of molecules: ligands, small molecules that bind to a receptor, and decoys, molecules with similar physical properties but dissimilar chemical structures compared to the ligands. Using DUD-E as a dataset for training, the performance of receptor-ligand and ligand-only ML models were compared to determine whether the receptor-ligand models also learned from the actual interactions or whether they just learned the inherent ligand bias. Ideally, the performance of these two types of models should be different as the binding energies are a function of both ligand and receptor structures, and the receptor structure can generally not be inferred from the ligand alone. However, the performance was found to be equivalent, the receptor-ligand models did not learn from the receptor structure, and it was concluded that "models may only learn the inherent bias in the dataset rather than physically meaningful features" \cite{chen2019hidden}. This calls for general methods to test what property predictors actually learn as it would naturally lead to more powerful, generalizable models and to more confidence in the models. To minimize hidden biases, the importance of dataset diversity for the generalizability of ML models has been pointed out before \cite{glavatskikh2019dataset}. \\

Finally, small datasets pose a particular challenge as they are inherently biased due to their limited size and, additionally, overfitting is to be expected. An important strategy to reduce both bias and overfitting in ML models when data is sparse is transfer learning (Figure 1-D) \cite{pan2009survey, cai2020transfer}. When using that approach, models are first pre-trained on large, sometimes unlabelled, datasets, and subsequently trained on the property of interest. Importantly, transfer learning has been shown to improve the prediction performance for models trained on small datasets significantly \cite{li2020inductive, goh2018using}. For instance, recently \cite{li2020inductive}, a language-based ML property prediction model was pre-trained first via self-supervised learning on one million unlabeled molecules curated from ChEMBL. This pre-training improved the prediction performance after subsequent training on small datasets of molecular activity consistently, regardless of the specific property prediction task and the training dataset size. Notably, the smallest dataset only contained 642 samples \cite{li2020inductive}.

\begin{figure*}[ht]
    \centering
    \includegraphics[width=0.92\textwidth]{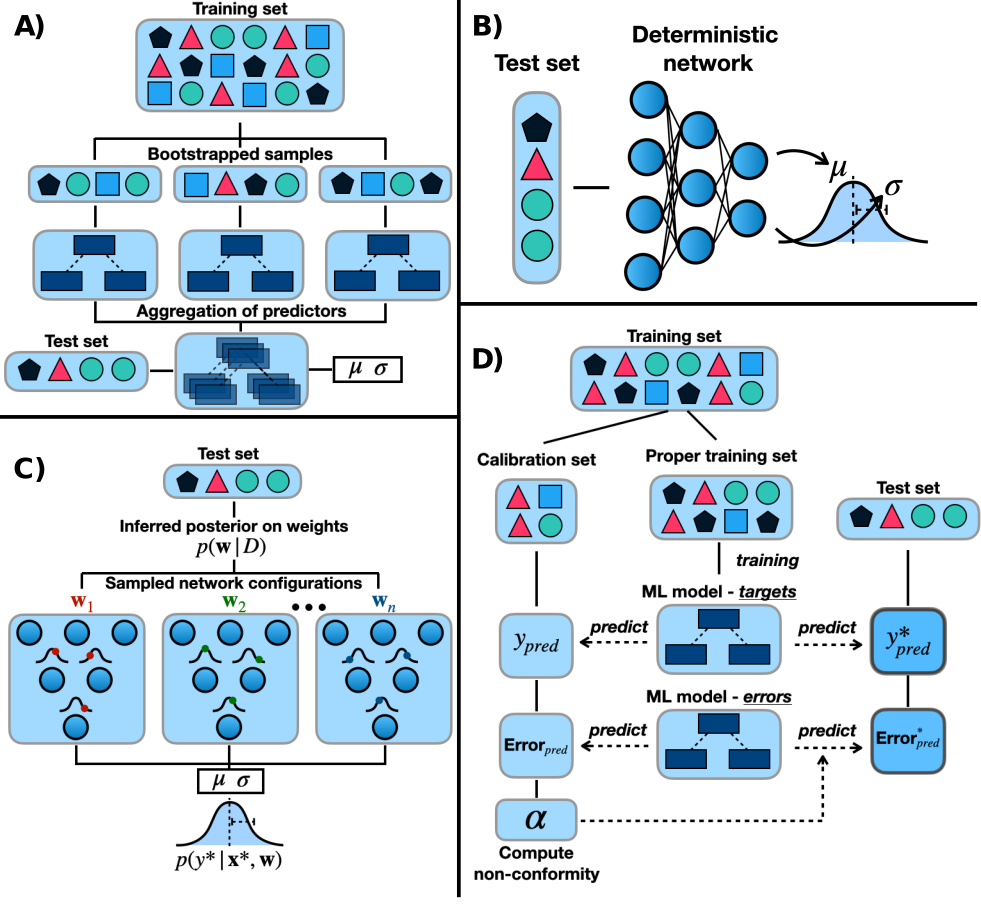}
    \caption{Four common methods for producing uncertainties or confidence intervals in molecular property prediction.  A) Bagging (bootstrap aggregation) algorithm with random forests as the ML model. B) Deterministic neural network trained using frequentist maximum likelihood estimation to approximate the distribution on the targets. C) Bayesian neural network in a regression setting depicting the test time sampling of different network configurations from the posterior distribution on the weights. D) Inductive conformal prediction in a regression setting using random forests as the ML model.}
    \label{fig:fig2}
\label{fig_x}
\end{figure*}

\section{Uncertainty in the Outputs}\label{chp:2}
Supervised ML largely relies on an inductive inference approach to derive general rules from a finite number of training examples. Therefore, predictive models obtained using inductive inference are never formally correct and inherently uncertain. There are often additional sources of uncertainty that stem from noise or imprecision on target measurements. For instance, molecular properties like binding affinity are often measured with an uncertainty, e.g. $1~\mu M \pm 10\%$. The latter type is referred to as aleatoric uncertainty, while the former is referred to as epistemic or systematic uncertainty and can in theory be reduced in light of more training observations or increased knowledge about the correct model. Both types of uncertainty are important, although they are often not treated separately and the formal distinction between the two remains an active area of research \cite{hullermeier2019aleatoric}. Many ML methods are available for representing uncertainty. They can be categorized with respect to the way uncertainty is represented and whether or not they allow for differentiation between the two types (aleatoric and epistemic). Here we divide the discussion of methods derived from frequentist statistics and ones based on Bayesian inference. \\

A widely used class of frequentist methods are ensemble strategies which typically use a large set of distinct predictors as opposed to a single model. Ensembles of tree-based models such as random forests \cite{breiman2001random} are commonly used to produce frequentist probability estimates based on the relative output frequency of the ensemble members \cite{sheridan2012three, toplak2014assessment}. Ensemble learning can introduce variance through multiple ensemble models with different parameter initializations (ensembling) or through the use of randomly sampled training sets (bagging or boosting, Figure 2-a). Ensembles of neural networks can also provide computationally efficient, readily parallelizable uncertainty estimates \cite{lakshminarayanan2016simple, scalia2020evaluating}. However, care must be taken using these methods as the uncertainties are often biased, poorly calibrated, and are only concerned with aleatoric uncertainty. Heteroscedastic aleatoric uncertainty can also be modeled using a single neural network trained with frequentist maximum likelihood inference that outputs a probability distribution (Figure 2-b) \cite{nix1994estimating}.  Conformal prediction (CP) has recently received significant attention, in which one ML model is trained to predict a property and another to predict the uncertainty (Figure 2 d) \cite{svensson2017modelling, norinder2014introducing, svensson2018maximizing, cortes2019concepts}.  CP methods are based on a rigorous mathematical framework and allow users to select intuitive confidence levels for their predictions, e.g. selecting a confidence level of 0.9 means that at most 10\% of predictions will be outside the predicted range. \\
 
Methods inspired by Bayesian inference seek to update a prior distribution $p(\theta)$ on the space of all possible models in light of training data, $D$, yielding a posterior distribution $p(\theta|D)$. Given a molecular representation for prediction, $x^*$, the predictive posterior distribution $p(y^*|x^*, D)$ is in theory obtained by averaging all possible models weighted by their respective posteriors. Bayesian methods have experienced a recent resurgence in drug design applications due to increasing computational power and advancement in algorithms for approximate inference. \\

Gaussian processes (GPs) \cite{rasmussen2003gaussian} allow for Bayesian inference in a non-parametric way, where the prior is determined by a mean and kernel function. The prior may be updated with observed data and for regression with Gaussian noise, the predictive posterior distribution is Gaussian with mean $\mu$ and variance $\sigma^2$. The predictive variance $\sigma^2$ represents the total uncertainty but can be decomposed into the variance of the error term $\sigma_\epsilon^2$ which represents aleatoric uncertainty. Epistemic uncertainty is the difference between $\sigma^2 - \sigma_\epsilon^2$ and is determined by the hyperparameters of the kernel function such as the characteristic length scale. GPs are commonly used tools for property prediction in the low data regime (i.e. below 1000 data points) due to their known robustness to overfitting \cite{rasmussen2003gaussian, costabal2019machine, bannan2018sampl6}. Hie et al. \cite{hie2020leveraging} used a GP trained on less than 100 compounds to screen a large library for candidates with nanomolar affinity for diverse kinases and whole-cell growth inhibition of Mycobacterium tuberculosis \cite{hie2020leveraging}. The authors show how uncertainty estimation can enable machine-guided discovery. However, the traditional implementation of GPs incurs cubic scaling with training set size, rendering them less desirable when training data is abundant. \\

Bayesian neural networks (BNNs) place probability distributions on neural network parameters $w$ and the analytically intractable posterior distribution $p(w|D)$ is typically estimated using variational Bayesian inference \cite{blundell2015weight}. The predictive posterior distribution is approximated using Monte Carlo sampling, which intuitively involves averaging the predictions of multiple parameter configurations sampled from the posterior (Figure 2-c). Replacing point-estimated network parameters with distributions allows for the quantification of epistemic uncertainty as well as heteroscedastic aleatoric uncertainty. Recently, such networks have been combined with state-of-the-art representation learning to give uncertain predictions of the physiological properties of small molecules \cite{zhang2019bayesian}.\\

For classification tasks, confidence calibration seeks to produce probability estimates which correctly reproduce the true correctness likelihood. In other words, given N predictions each with confidence 0.9 about a label, we would expect 90\% of the predictions to give the correct label. Scalar valued confidence statistics such as expected calibration error and maximum calibration error have also been proposed. Many calibration methods such as isotonic regression or temperature scaling are based on a learned post-processing step in which a calibrated probability $q_i$ is derived from the model’s output probability $p_{i^*}$ \cite{guo2017calibration}. Confidence calibration schemes can be applied as a scalable post-processing step to any predictive learning algorithm, and their ideas can be extended to regression tasks. \\

Confidence or uncertainty estimation in molecular property prediction is crucial for tasks that involve data-driven decision making where valuable resources or human risk is at stake. For example, BNNs have been employed to give intuitive probabilistic statements about the severity of the risk of drug-induced liver injury \cite{williams2019predicting, semenova2020bayesian}. Predicting clinical outcomes from pre-clinical data and using uncertainty estimates to quantify and communicate risks can preserve time, resources, and patient well-being \cite{lazic2018predicting}. Predictive uncertainty is central to the field of active learning \cite{settles2009active}, which has been known in the context of drug discovery for almost two decades. Active learning strategies balance exploitation of the knowledge of a predictive model to retrieve active candidate compounds with the exploration of regions in compound space where the model is highly speculative \cite{zhang2019bayesian, reker2015active, reker2017active}. Explorative behaviour of an active learner is paramount when structural novelty is prioritized. Elucidation of error sources in training data remains a critical application for uncertain predictive models. Ryu et al. modeled aleatoric and epistemic uncertainty separately using Bayesian graph convolutional neural networks for a variety of molecular property prediction tasks including classification of bioactivity and toxicity. Analysis of predictive aleatoric uncertainty allowed for the identification of poor quality training data for some molecular candidates \cite{ryu2019bayesian}. \\

In the near future, we believe predictive uncertainty will become an indispensable component of molecular property prediction. Researchers will continue to systematically benchmark uncertainty quantification methods on various datasets \cite{hirschfeld2020uncertainty}. We anticipate software packages that facilitate uncertain predictions will continue to appear \cite{moss2020gaussian}. \\

\begin{figure*}[ht]
    \centering
    \includegraphics[width=0.92\textwidth]{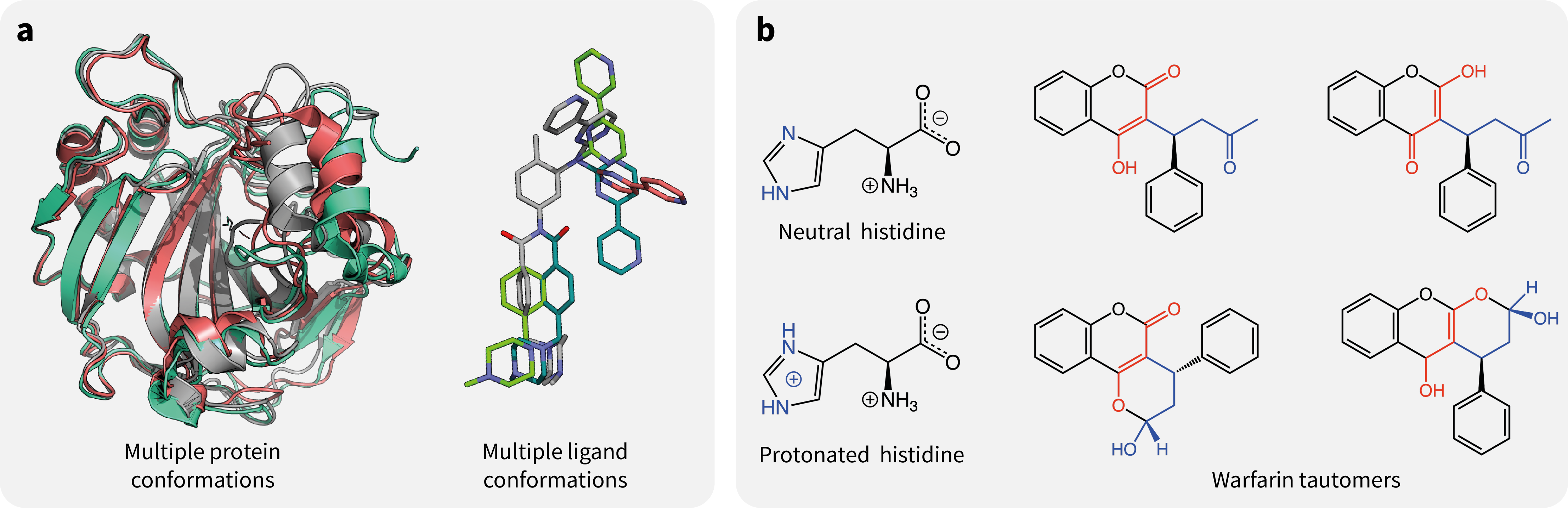}
    \caption{Examples of uncertainty in the structural and chemical inputs used by ML models. (a) Conformational variability is encountered both in proteins and small organic molecules. On the left-hand side, some of the conformations of carbonic anhydrase II are shown, as observed via nuclear magnetic resonance (PDB-ID 6HD2). On the right-hand side are some of the possible conformations of Imatinib (computed with Frog \cite{frog_}). (b) Examples of major protonation and tautomeric states present in the solution for an amino acid and a drug. The protonated form of histidine has a $pK_a$ of approximately 6.0, which means that it is generally expected to be found in its neutral form at physiological pH. Yet, because of the local environment in protein binding pockets, it is often unclear which of the two is the major species, and they may both co-exist in non-negligible proportions. On the right-hand side, some of the tautomeric states of warfarin \cite{guasch2015tautomerism} are shown.}
    \label{fig:overview}
\label{fig_x}
\end{figure*}

\section{Uncertainty in the Input Features}\label{chp:3}
Contrary to uncertainty in the properties being predicted, imprecision in the input features is a source of uncertainty that is rarely considered. Property prediction in drug design often relies on representations encoding a specific molecule, such as SMILES strings \cite{weininger1988smiles}, or on three-dimensional structural data. However, these static descriptions of molecular systems are not representative of their dynamic nature in solution. In this section, we discuss how input-related uncertainty might arise when predicting chemical properties from molecular structures, and how some of the latest ML models might be used to capture this uncertainty in the confidence of the predictions. \\

One form of input uncertainty is due to the presence of multiple protonation and tautomeric states in solution. String and graph representations of molecules are static and unable to capture these multiple states and their probabilities in solution. SMILES strings representing two different tautomers are likely to return different property predictions when they in fact should be the same given the behaviour of these molecules in solution. This represents a challenge to many compound property predictions and cheminformatics tools and may be seen as a form of input uncertainty, given that the chemical species in solution can be modeled only approximately. Similarly, protonation states depend on the chemical environment of the molecule, there may not be one dominant state in solution and, finally, these may change upon interaction with a binding partner. All these aspects contribute to creating uncertainty about the input representation used in ML prediction tasks. \\

Three-dimensional structural data is often used to predict properties like drug binding affinity, selectivity, and mutational effects on protein stability and drug resistance. This data is typically derived by X-ray crystallography and, increasingly, cryo-electron microscopy (cryo-EM). Despite providing abundant information on protein-ligand interactions, which can be used to drive drug design, these structures are only a single snapshot of all possible protein-ligand conformations in solution. As such, this three-dimensional data is a single datapoint of a much broader input probability distribution. Some structure-based modeling approaches, like ensemble docking, try to take this uncertainty into account by sampling multiple conformations with molecular dynamics simulations. In such a way, ensemble docking tries to capture the distribution/uncertainty of protein-ligand structures. By generating an “ensemble” of drug target confirmations, the docking results are less likely to be overfitted to a specific protein conformation. However, fast scoring functions for ligand binding affinities are typically trained on databases of static structures, such as the PDBbind database, and do not capture this type of uncertainty in their predictions. In addition to the statistical uncertainty caused by conformational variability, differences between protein structures \textit{in crystallo} and \textit{in vivo} due to packing effects and cryo temperatures may result in systematic biases.\\

Ideally, one would like uncertainty about multiple possible chemical and conformational states in solution to be captured by the ML model and reflected in the confidence of the property predictions. Currently, only a handful of specialized ML approaches are able to capture uncertainty in the input features. Gaussian process latent variable models, together with other similar kernel-based approaches \cite{girard2003gaussian, hanafusa2020bayesian}, are examples of such models \cite{titsias2010bayesian, li2016review}. Recent work has also started to investigate the effect of molecular ensembles in property predictions \cite{axelrod2020molecular} with ML models. However, we are unaware of widely adopted supervised ML models that can either take into account known input uncertainty, or that can infer it directly from data. We believe this to be an underexplored research direction with much potential for practical applications. Such models could become an indispensable component of molecular property prediction, being able to capture or infer feature uncertainty in structural and chemical data.

\begin{figure*}[ht]
    \centering
    \includegraphics[width=0.90\textwidth]{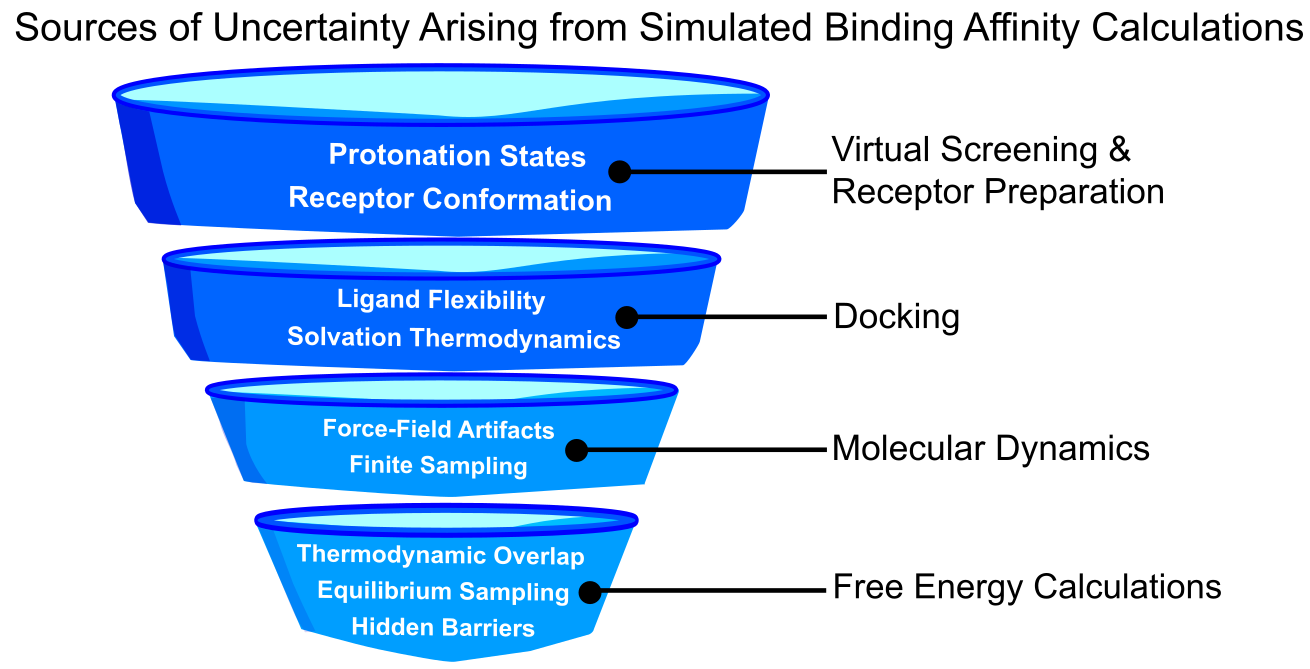}
    \caption{A generalized sequence of steps to compute binding affinities, and their corresponding facets of uncertainty. The top half of the diagram illustrates low-cost methods, where the sources of uncertainty can be more easily identified. The bottom half of the pathway refers to the methods which sacrifice computational cost for accuracy, and for which the minimization of uncertainty largely represents open problems in computing accurate binding affinities.}
    \label{fig:overview}
\label{fig_x}
\end{figure*}

\section{Uncertainty in the Binding Affinities}
A key property of interest in drug discovery is the binding affinity of a ligand to a target receptor. Of the many physics-based models that have been developed and refined to predict binding affinity, here we focus on two widely-used techniques that utilize the three-dimensional structures of ligand and receptor, and a molecular-mechanics energy function to compute ligand affinities: molecular docking and free energy perturbation (FEP) methods.  \\

Of the two, molecular docking is less computationally expensive and more suitable for high-throughput virtual screening of potential inhibitors. Docking algorithms such as AutoDock Vina \cite{trott2010autodock}, Glide \cite{friesner2004glide}, Surflex-Dock \cite{spitzer2012surflex} and GOLD \cite{verdonk2003improved} can predict correct bound poses ($<2$\r{A} RMSD) with an accuracy of 70-80\% \cite{cheng2009comparative} and can correctly rank (trios of low-, medium- and high-affinity) ligands with an accuracy of almost 60\% \cite{cheng2009comparative, li2014comparative, khamis2015machine}. Predictions of absolute binding affinity from docking scores, however, generally show a poor correlation to experimental binding affinities \cite{procacci2021methodological}, with pK\textsubscript{a} (decadic logarithm of the affinity) prediction accuracies of 1.5-2.0 (RMSE) \cite{li2014comparative, jimenez2018k}. \\

Large uncertainties in docking estimates can be attributed to conceptual deficiencies in the scoring function, for example, a lack of proper accounting of solvation thermodynamics, which may require incorporation of explicit water molecules against an ensemble of receptor poses to better represent the physical system \cite{pantsar2018binding}. Docking scores are also made uncertain by implementation choices made for algorithmic efficiency. For example, flexible ligand or receptor degrees of freedom are usually restricted to enable efficient conformational searching.  Recent studies using deep learning neural-network-based scoring functions show promise in reducing prediction uncertainties \cite{jimenez2018k, morrone2020combining}. Regardless of these inaccuracies, docking approaches are likely to remain an essential component of virtual screening, as they can reduce and enrich the chemical space of potential binders, and provide initial bound-pose conformations for seeding molecular dynamics (MD) simulations. \\

In contrast to docking, FEP methods are much more expensive, requiring all-atom MD simulations to perform statistical sampling of ligand-receptor configurations \cite{shirts2007alchemical, shirts2013introduction, gapsys2015calculation, cournia2020rigorous}.  The benefit is a best-in-class prediction of ligand binding affinity, achieving accuracies within chemical accuracy, i.e. 1.0 kcal/mol \cite{wang2015accurate, aldeghi2016accurate}. Assigning uncertainties to such predictions, however, requires consideration of the many error sources these methods introduce. These can be categorized as arising from both sampling and scoring inaccuracies. \\

\subsection{Uncertainty in FEP Estimates due to Sampling Error} 
To understand uncertainty from statistical sampling error, consider that using MD simulation naïvely to predict ligand affinity would require equilibrium sampling of multiple reversible binding transitions which typically exceed the timescales of state-of-the-art MD, even on special-purpose high-performance computers \cite{shan2011does}. FEP and related methods circumvent this problem by performing simulations in multiple thermodynamic ensembles for a series of alchemical intermediates and analyzing the resulting samples with a statistical free energy estimator. Absolute binding free energy (ABFE) is computed along alchemical paths that decouple all non-bonded interactions in solution of the ligand, including the binding to a receptor.  Relative binding free energy (RBFE) is computed along with an alchemical transformation from one molecular topology to another.\\

Statistical free energy estimators that utilize multi-ensemble sampling include exponential averaging (EXP) \cite{zwanzig1954high}, thermodynamic integration (TI) \cite{kirkwood1935statistical}, the weighted-histogram method (WHAM) \cite{kumar1992weighted}, the Bennett acceptance ratio (BAR) method \cite{bennett1976efficient}, the multistate Bennett acceptance ratio (MBAR) method \cite{shirts2008statistically}, and newer dynamical estimators like TRAM \cite{wu2016multiensemble} and DHAM \cite{rosta2015free}. Uncertainties from statistical sampling can influence the confidence of calculated Gibbs free energies in complicated ways. Hence, regardless of which free estimator is used, a non-parametric bootstrap is preferred to estimate uncertainty in Gibbs free energies \cite{shirts2013introduction}.  Similarly, uncertainties can be calculated over block-averages of a long and equilibrated simulation trajectory, or can alternatively be estimated across multiple independent simulations, in some cases initiated with different ligand starting poses. For FEP, which involves computing free energy differences between the ligand-bound state and the free ligand in solution, overall uncertainties are often obtained by summing individual uncertainties for each ensemble in quadrature \cite{procacci2021methodological, shirts2013introduction}. \\

Different free energy estimators have unique bias and variance due to finite sampling, but they all depend on having sufficient thermodynamic overlap between neighbouring ensembles along the thermodynamic path. For this reason, relative free energy estimation is often more accurate for pairs of similar molecules, since the alchemical transformation involves a smaller number of atoms, and hence inherently has a better overlap. Achieving good thermodynamic overlap is the main consideration guiding choices about the number and type of alchemical intermediates, which may require position or angular restraints on the ligand to focus sampling on particular ligand poses \cite{procacci2021methodological}. \\

Most free-energy estimators assume that samples are statistically independent and drawn from the equilibrium distribution. Sufficient equilibration can be notoriously slow to achieve for some systems, e.g. ligands with “hidden barriers” between conformational states, a problem which Hamiltonian replica exchange and other enhanced sampling methods have been designed to address \cite{wang2012pubchem}. To avoid statistically correlated input, which results in artificially low uncertainty estimates, recommended strategies discard initial unequilibrated samples, and subsample snapshots at a frequency $(2\tau+1)$, where $\tau$ is a computed correlation time \cite{grossfield2018best}.  An algorithm to automatically detect simulation equilibration in this way has been developed \cite{chodera2016simple}. \\

Related to FEP is a class of non-equilibrium work (NEW) free energy methods, based on the Crooks fluctuation theorem \cite{crooks1999entropy}, which drives a system between alchemical endpoints and analyzes the distribution of non-equilibrium work values \cite{gapsys2015calculation}.  These methods can achieve comparable accuracies to FEP with fewer intermediates \cite{gapsys2020large, baumann2020challenges, khalak2020non}, yielding uncertainties that depend on the variance of the work distributions, in turn depending on the timescale of the alchemical schedule. Sampling uncertainty can also be introduced by artifacts from the implementation of MD.  For example, uncertainties on thermodynamic and kinetic properties can depend greatly on solvation artifacts that depend on simulation box size \cite{gapsys2020importance}. Furthermore, surprisingly, variation in computed energies across different simulation packages \cite{shirts2017lessons}, integrators, and thermostats may be non-negligible \cite{merz2018testing}. \\

\subsection{Uncertainty in FEP Estimates due to Scoring Error}
The workhorse of FEP is a molecular mechanics energy function, called a force field, which has usually been designed to approximate a quantum mechanical potential energy surface, and tuned to reproduce experimental observables. Iterative cycles of force field development, parameterization and assessment are thus vital for reducing the uncertainty introduced by force field inaccuracy \cite{khalak2020non, beauchamp2012protein, nerenberg2018new}. With new machine learning advances have come new ways to improve force fields for the prediction of ligand binding free energies.  One product of the recently-launched Open Force Field Initiative is a new generation of force fields that eschews atom-types in favour of direct chemical perception, utilizing the SMIRKS-native Open Force Field (SMIRNOFF) scheme for parameter assignment \cite{mobley2018escaping}. Parameterization of Open Force Field v1.0.0, code-named Parsley \cite{svensson2018maximizing}, was made possible by the ForceBalance algorithm, a Bayesian algorithm for automatic and reproducible parameterization \cite{wang2014building, wang2017building, check}. Polarizable force fields such as AMOEBA continue to make gains in accuracy for predictions of Gibbs free energy of binding \cite{ponder2010current, shi2020amoeba}, and there are examples of neural network-based potentials, which aim to reproduce molecular properties with DFT accuracy at force field cost \cite{smith2017ani, lahey2020simulating}. \\

Finally, uncertainty can also result from the assumptions of a selected model.  For example, most fixed-charge force fields require the pre-selection of particular protonation states for both ligand and ionizable side chains of a receptor. Depending on the predicted pK\textsubscript{a} of acidic or basic groups, which may differ significantly between bound and unbound states, multiple simulations may be required to explore these possibilities. Overall, there are several factors to consider when designing a study to compute binding affinities for drug-like molecules. New methodologies and best practices can often be found through blind assessment challenges such as SAMPL \cite{icsik2020assessing} (Statistical Assessment of Modeling of Proteins and Ligands), which are created annually to assess predictions of binding affinities among other properties important to drug discovery. \\

\begin{figure*}[ht]
    \centering
    \includegraphics[width=1.00\textwidth]{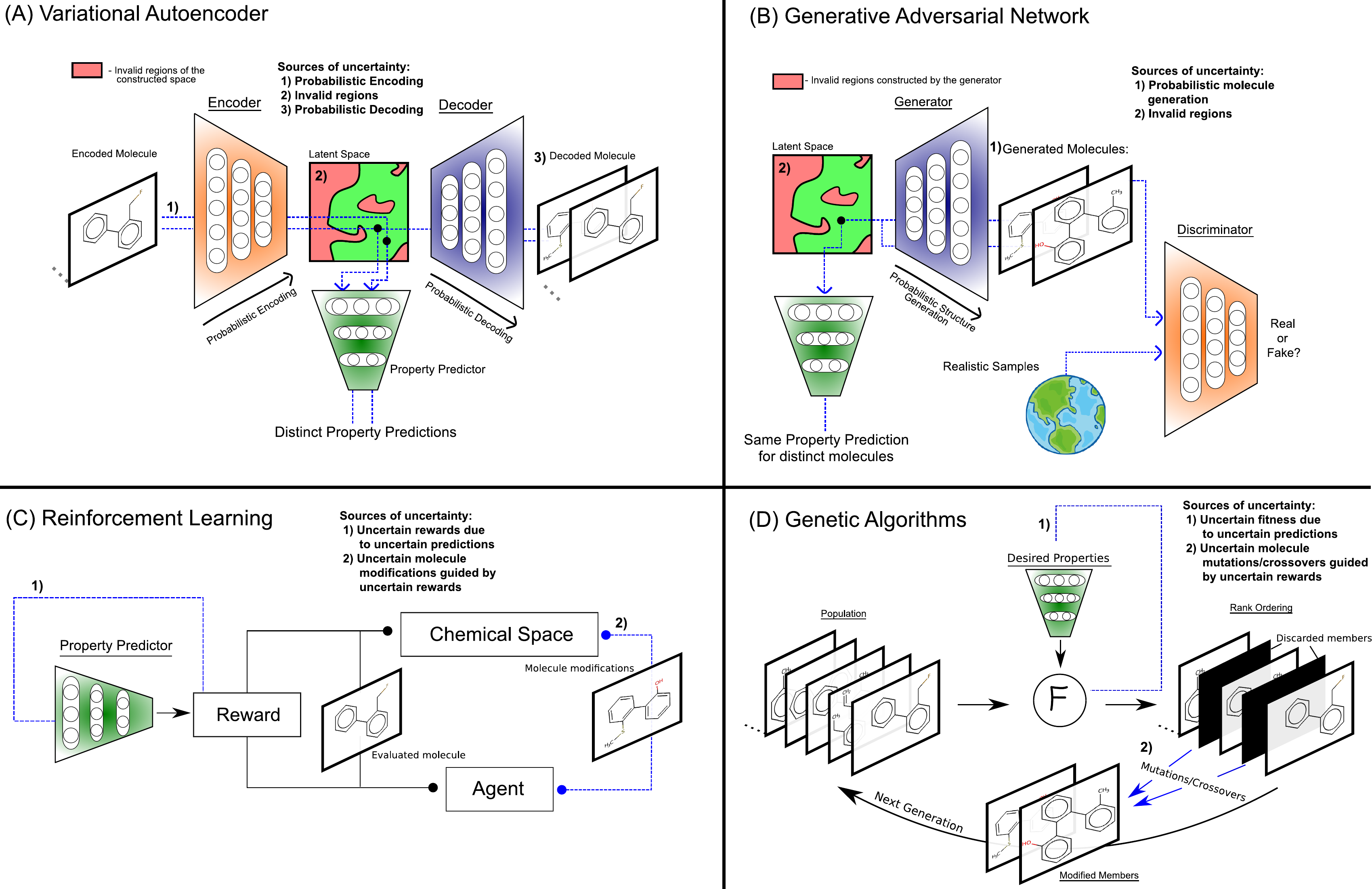}
    \caption{ Architectures and sources of uncertainty of generative models for molecular design, namely: (A) Variational autoencoder, (B) Generative adversarial network, (C) Reinforcement learning, and (D) Genetic algorithms. Sources of uncertainty within each architecture are highlighted in blue.}
    \label{fig:overview}
\label{fig_x}
\end{figure*}

\section{Uncertainty in Generative Modeling}
Generative models can produce synthetic data samples, some of them by learning from reference data. For molecular design, generative models are used to tackle the inverse design problem \cite{sanchez2018inverse}, where novel structures with desired properties are generated. Molecular property estimators are often used in conjunction with these models \cite{pollice2021data}.  In this section, we discuss the importance of property prediction uncertainty in the context of generative modeling. In particular, we consider variational autoencoders (VAEs) \cite{kingma2013auto}, generative adversarial networks (GANs) \cite{goodfellow2014generative}, reinforcement learning (RL) \cite{sutton2018reinforcement}, and genetic algorithms (GAs) \cite{mitchell1998introduction}. \\

VAEs possess the ability to take discrete structures such as molecules and convert them into high-dimensional latent representations by way of neural networks (referred to as encoders). Simultaneously, a mapping from the latent representation back to the discrete structure is trained using a separate neural network (referred to as a decoder). During training, VAEs learn to encode and decode correct molecules and simultaneously shape the latent space into a gaussian distribution for sampling. For molecular design, Gómez-Bombarelli et al. \cite{gomez2018automatic} introduced an additional neural network that predicts molecular properties from the latent space representation. During the training of this network, gradient optimization steps are taken for both the parameters of the encoder network and the property prediction network. This shapes the latent space based on the predicted property. \\

The formulation of VAEs as generative models brings about three sources of uncertainty for the associated property prediction model: (1) Decoding points from a latent space does not always lead to valid molecules. However, for every point in the latent space one can predict molecular properties via the prediction model. Hence, these regions can be viewed as uncertain spaces where predictions have to be disregarded due to the lack of a physical manifestation. (2) The encoding of structures to the latent space is stochastic. Kingma et al. \cite{kingma2013auto} introduced a reparameterization step that ensures differentiability through the latent space for training the overall architecture and imposition of a gaussian distribution. As a result, identical molecules are typically encoded at a range of points in the latent space. While these encoded molecules should have the exact same predicted property, the range of positions tends to give rise to different predictions. Hence, there are multiple property predictions associated with any given molecule representing the corresponding uncertainty. (3) Discrete objects within VAEs, in particular molecular string representations and adjacency matrices, are usually decoded from the latent space as a time-series. These discrete objects are generated one at a time, by sampling over the distribution of all user-defined actions, i.e. units such as the allowed characters in a string. Typically, the unit with the largest probability is selected. However, multiple units can have identical probabilities. Consequently, decoding the same point from the latent space multiple times can result in distinct molecules. Additionally, as the same latent space point represents all these different molecules simultaneously, they will have the exact same predicted property. \\

GANs \cite{goodfellow2014generative, guimaraes2017objective, sanchez2017optimizing} are generative models with joint training of two competing networks, a generator and a discriminator. Like VAEs, they rely on a gaussian latent space to draw samples from and decode them using a neural network (referred to as generator network). The generated samples are compared to references from a dataset by another neural network (referred to as discriminator). The generator and discriminator compete with each other, the generator tries to produce samples that appear more realistic to the discriminator while the discriminator tries to improve its reliability to distinguish generated from reference samples. Similar to VAEs, in the traditional formulation of GANs, property predictors can be associated with the latent space. The sources of uncertainty are essentially identical to VAEs. \\

RL is a field where agents learn to navigate environments to maximize rewards. In molecular design, RL agents build chemical compounds \cite{staahl2019deep, zhou2019optimization, popova2018deep}, and rewards are provided for achieving desired molecular properties. Actions constitute making changes to chemical structures, while rewards are calculated on-the-fly based on the resulting structures. In contrast, GAs \cite{nigam2019augmenting, jensen2019graph, yoshikawa2018population} are generative models that generate structures without gradient-based optimization. The algorithm is initiated with a set of molecules, i.e. the initial population, that are ranked by calculating their fitness. For molecular design, the fitness is typically assessed based on a set of target properties. The fittest molecules are carried over to the next generation, the unfit molecules are replaced with mutations and crossovers of the best. For both RL and GAs, to offset the often prohibitive computational cost of obtaining accurate property predictions, cheap approximations such as ML models are utilized to estimate rewards and fitness, respectively. By using cheap property predictors, in high-dimensional spaces, the generation of structures can be skewed to uncertain regions of the chemical space. For instance, certain structures can be adversarial examples or outliers for property predictors, where the model can be overly confident or uncertain about its prediction. Hence, the resulting molecule modifications are uncertain due to the uncertainty in the rewards. Notably, Thiede et al. \cite{thiede2020curiosity} demonstrated that the uncertainty from prediction networks can be used directly in the reward function to guide the generation of structures outside of the confidence region of the prediction network for curiosity-driven molecular design. Interestingly, Shen et al. \cite{shen2020deep} propose PASITHEA, where a property prediction neural network is directly used for generating molecules with deep dreaming. An interesting direction would be the incorporation of uncertainty within property predictors for structure generation via the deep dreaming algorithm.\\

Generative models coupled with property predictors are yet to achieve wide adoption due to their complex setup, their high computational costs and the need to adapt them to specific problems. When used for properties that are inherently complicated and hard to predict, the role of uncertainty and prediction confidence will be very important. In addition, in models with latent spaces such as VAEs and GANs, the choice of molecular representation influences both the validity of the latent space and thereby the prediction confidence. For instance, the use of VAEs and GANs was demonstrated in conjunction with the SMILES \cite{weininger1988smiles} representation, where the probability to produce invalid strings is high. Consequently, both VAEs and GANs trained on SMILES can produce invalid molecules. Importantly, this problem has recently been solved by the SELFIES representation \cite{krenn2020self}, which guarantees molecular validity by design. While SELFIES increases model confidence by generating only valid molecules, it does not remove uncertainty from inherent model stochasticities such as probabilistic encoding and decoding. Furthermore, confidence in molecular property predictors can be increased by extending datasets used for training and validation of property predictors systematically. Recently, STONED \cite{nigam2020beyond}, a simple generative model producing molecules at high speed via systematic navigation within SELFIES, has been introduced. Accordingly, based on an initial dataset, it can be used to diversify the molecule space systematically leading to better property prediction models.

\section{Conclusion}
In this review, we inspected common sources of uncertainties encountered in data-driven drug design workflows, some of which are well-studied in the literature, and some of which are typically neglected. Hidden biases in data can lead to overly confident property predictions that fail to manifest in the real world. Data-driven prediction models that explicitly incorporate uncertainty convey insights into the confidence of the model itself and provide a handle to assess the prediction quality. Accounting for uncertainty in the input features is underappreciated in ML research but is inherent to complex systems such as the chemistry of the human body. In contrast, error estimation in the prediction of binding free energies is one of the central methods in the field to assess the quality of simulation workflows that guides the development of efficient sampling techniques. Uncertainties in generative models are another underappreciated topic that needs to be addressed to improve the success of property-driven molecular design approaches. Overall, it is evident that assigning confidence in molecular property predictions is intimately tied to understanding the associated sources of error, and we believe that future progress in this field will require the development and application of even more methods accounting for uncertainties.

\section{Expert Opinion}
Assigning confidence to predictions is directly connected to uncertainty identification and quantification. When training ML models for predicting complex molecular properties, users need to be aware of potential biases of the training data when assessing generalizability. The recent use of DUD-E to train data-driven models serves as a cautionary tale for models learning to exploit biases in the data rather than learning the underlying physics of intermolecular interactions \cite{mysinger2012directory}. Most importantly, the smaller the training dataset size the larger the danger of hidden bias and the higher the importance of accounting for them. In that regard, transfer learning, especially in extremely sparse data regimes, requires more research to establish it as the method of choice for molecular property prediction. \\

Moreover, not only do small datasets lead to significant biases but they also lead to higher epistemic uncertainty because the model does not have enough information to learn the true relationship between input features and output labels \cite{hullermeier2019aleatoric}. However, expanding datasets, especially for properties relevant in drug design, can be prohibitively expensive and very time-consuming, in particular in the case of experimental data. Furthermore, not every new datum added to an existing dataset provides the same amount of additional information. Accordingly, developing systematic methods to identify data to be added to the existing dataset to maximize the improvement in prediction accuracy and the reduction in epistemic uncertainty is the subject of ongoing research. \\

Additionally, the aleatoric uncertainty  \cite{hullermeier2019aleatoric} that is inherent in the data used to train ML models is at least as important as the epistemic one. One prominent example from the field of drug discovery discussed in this review is the use of estimated and simulated binding affinities  to assess the interaction strength between ligands and receptors. Notably, binding affinities estimated via molecular docking are only of limited accuracy, but their computation efficiency allows assembling large datasets. The main question is whether the estimated property, regardless of how efficient it can be computed, conveys any information about the real relationship between input features and model output. When noise dominates over valuable information, the largest dataset will not provide useful information for ML models to be learned. In summary, it is important to build datasets and construct models that explicitly and consistently account for both aleatoric and epistemic uncertainties. While uncertainty estimates are relatively straightforward to obtain in general, they are only rarely benchmarked or evaluated during model training and future work needs to establish robust procedures for that purpose.\\

Another important and often underappreciated source of uncertainty, especially in complex systems, are the input features.  This is particularly true for chemistry where molecular properties are generally conformer-dependent, and most molecules adopt much more than just one conformation under ambient conditions. Data-driven models, in contrast, are largely trained on 2D molecular representations that neglect conformations entirely. Accordingly, incorporating 3D structures and appropriate features in ML models is an active field of research \cite{axelrod2020molecular}. The main challenge is to offset the added cost of generating representative 3D conformations with higher prediction accuracy in the final models. Currently, 2D molecular representations are still the state of the art in the field but we believe 3D features will become the standard in the near future. \\

One topic of paramount importance in drug design is simulating binding affinity between ligands and receptors. For that purpose, docking is a popular technique that provides crude estimates of interaction geometries and strengths using extremely efficient scoring approaches. However, this efficiency comes at the price of accuracy and, hence, cannot match the quality of FEP simulations, which are several orders of magnitude more computationally demanding. Importantly, even among FEP approaches there is an inevitable trade-off between cost and accuracy as the extent of sampling conducted directly controls the quality of the results. Accordingly, for extremely accurate results \cite{procacci2021methodological}, FEP methods need to capture multiple binding transitions and this can increase the simulation cost by orders of magnitude making it less practical.  Finally, a large portion of the uncertainties from both docking approaches and MD simulations stem from the scoring functions and the force fields used, respectively. While systematic optimization procedures for force field parameters are being developed and implemented, the development of force fields that not only provide energy but also uncertainty estimates has received less attention and we believe that future work in that direction is necessary to deliver more accurate models. \\

Finally, the generative models for molecular design introduced recently rely on efficient molecular property predictors to generate molecules with both desirable structure and function \cite{pollice2021data}. However, an explicit account and treatment of uncertainties, unfortunately, is largely neglected in the field. While accounting for property prediction uncertainties seems relatively straightforward with current methodologies, dealing with uncertainties in the molecular structure itself, and modifications thereof remains unexplored territory in comparison. Accordingly, we believe that future research in that direction is paramount and has the potential to lead to significant advances in computer-driven molecular design in general and drug design in particular.

\section*{Acknowledgements}
R.P. acknowledges funding through a Postdoc.Mobility fellowship by the Swiss National Science Foundation (SNSF, Project No. 191127). R.J.H. gratefully acknowledges NSERC for provision of the Postgraduate Scholarships-Doctoral Program (PGSD3-534584-2019). M.A. is supported by a Postdoctoral Fellowship of the Vector Institute. V.A.V and M.F.D.H acknowledge support in part by National Institutes of Health grant 1R01GM123296. A. A.-G. thanks Anders G. Frøseth for his generous support. A. A.-G. also acknowledges the generous support of Natural Resources Canada and the Canada 150 Research Chairs program. We also acknowledge the Department of Navy award (N00014-19-1-2134) issued by the Office of Naval Research. The United States Government has a royalty-free license throughout the world in all copyrightable material contained herein. Any opinions, findings, and conclusions or recommendations expressed in this material are those of the authors and do not necessarily reflect the views of the Office of Naval Research.

\begin{figure*}[h]
    \caption{TOC Image}
    \centering
    \includegraphics[width=0.80\textwidth]{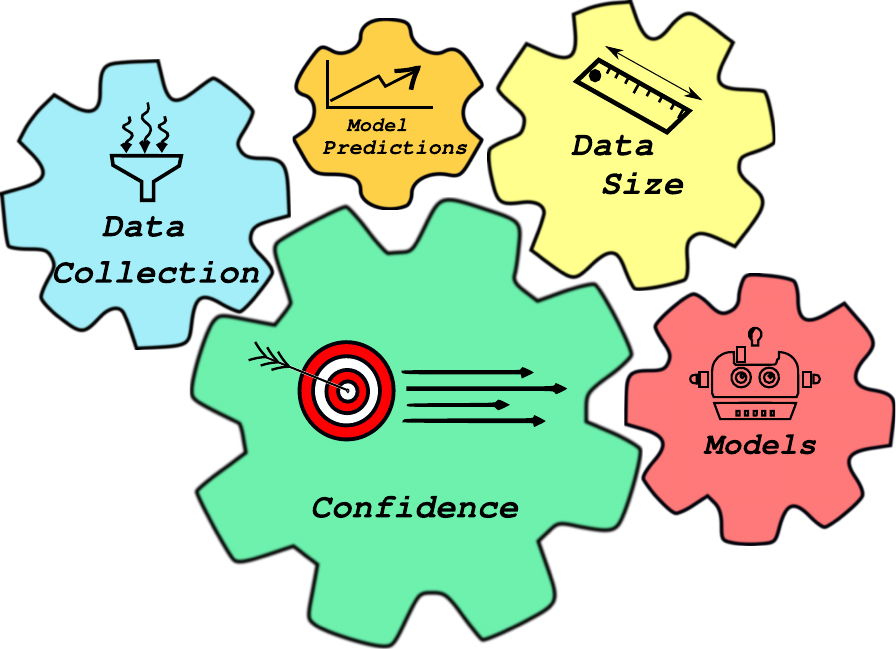}
    \label{fig:TOC}
\label{fig_x}
\end{figure*}

\bibliographystyle{unsrt}
\bibliography{refs}

\clearpage
\newpage
\widetext

\setcounter{section}{0}
\setcounter{figure}{0}
\renewcommand{\thepage}{S\arabic{page}} 
\renewcommand{\thesection}{S\arabic{section}}  
\renewcommand{\thetable}{S\arabic{table}}  
\renewcommand{\thefigure}{S\arabic{figure}}

\end{document}